\documentclass[times, review, 10pt]{elsarticle}

\usepackage[numbers]{natbib}
\usepackage[utf8]{inputenc}
\usepackage{url}
\usepackage{booktabs}
\usepackage{amsfonts}
\usepackage{nicefrac}
\usepackage{microtype}
\usepackage{xcolor}
\usepackage{graphicx}
\usepackage{adjustbox}
\usepackage{amsmath}
\usepackage{multirow}
\usepackage[inline]{enumitem}
\usepackage{placeins}
\usepackage{balance}
\usepackage[
    unicode,
    hidelinks,
    pdftitle={Can We Use Saliency Maps for the Evaluation of Generative Image Models?},
    pdfauthor={Eleonora Ristori, Luca Bindini, Marco Lippi, Paolo Frasconi}
]{hyperref}

\definecolor{myblue}{HTML}{2064AA}
\definecolor{myred}{HTML}{AE172A}
\DeclareMathOperator*{\median}{median}

\begin{document}

\begin{frontmatter}
\title{Can We Use Saliency Maps for the Evaluation of Generative Image Models?}

\author[1]{Eleonora Ristori\corref{corresponding}}
\ead{eleonora.ristori@unifi.it}

\author[1]{Luca Bindini}
\ead{luca.bindini@unifi.it}

\author[1]{Marco Lippi}
\ead{marco.lippi@unifi.it}

\author[1]{Paolo Frasconi}
\ead{paolo.frasconi@unifi.it}

\address[1]{AI Lab, DINFO, University of Florence, Florence, Italy}

\cortext[corresponding]{Corresponding author.}

\begin{abstract}
Evaluating generative image models remains challenging, as commonly used metrics depend on fixed pretrained feature extractors optimized for different tasks that do not always align with human realism perception across datasets and model families. To address this limitation, we investigate a human-inspired evaluation framework that avoids representations learned for unrelated tasks. Specifically, we train a generator-specific classifier to distinguish real images from samples produced by a given model and analyze its decisions using saliency maps. Across diverse benchmarks, we observe that these classifiers learn generator-specific artifacts rather than a universal notion of “fakeness”, exhibiting strong accuracy and limited cross-generator transfer. This makes saliency analysis particularly informative: the resulting attribution maps highlight visual cues and structural irregularities characteristic of each model’s outputs. By combining classification accuracy and saliency information, we obtain a composite score aligned with human judgments, without relying on large pretrained backbones and using only real and generated samples from the evaluated model.
\end{abstract}


\end{frontmatter}

% \maketitle

\section{Introduction}
\label{sec:intro}

In recent years, generative image models have achieved remarkable progress,
producing synthetic images that are often hard to distinguish from real
ones~\cite{DBLP:conf/nips/DhariwalN21}. This
progress has been driven by several families of generative models, including
explicit-likelihood models such as variational autoencoders
(VAEs)\cite{DBLP:conf/nips/VahdatK20, DBLP:journals/corr/abs-2203-13751}, adversarial models such as
GANs~\cite{DBLP:conf/nips/GoodfellowPMXWOCB14} and their large-scale variants
(e.g., BigGAN~\cite{DBLP:conf/iclr/BrockDS19} and
StyleGAN/StyleGAN-XL~\cite{DBLP:conf/cvpr/KarrasLA19,DBLP:conf/siggraph/SauerS022}),
as well as diffusion and score-based models
(DDPM~\cite{DBLP:conf/nips/HoJA20}, LDM~\cite{DBLP:conf/cvpr/RombachBLEO22}),
which currently provide strong fidelity--coverage trade-offs. Despite these
advances, evaluation methodologies have not progressed at the same pace.
Ideally, evaluation should rely on human studies, which are, however,
expensive, difficult to scale, and sensitive to experimental design, thus
motivating research towards the definition of algorithmic evaluation metrics
that are aligned with human judgment.

\begin{figure}[t]
    \centering
    \includegraphics[width=\linewidth]{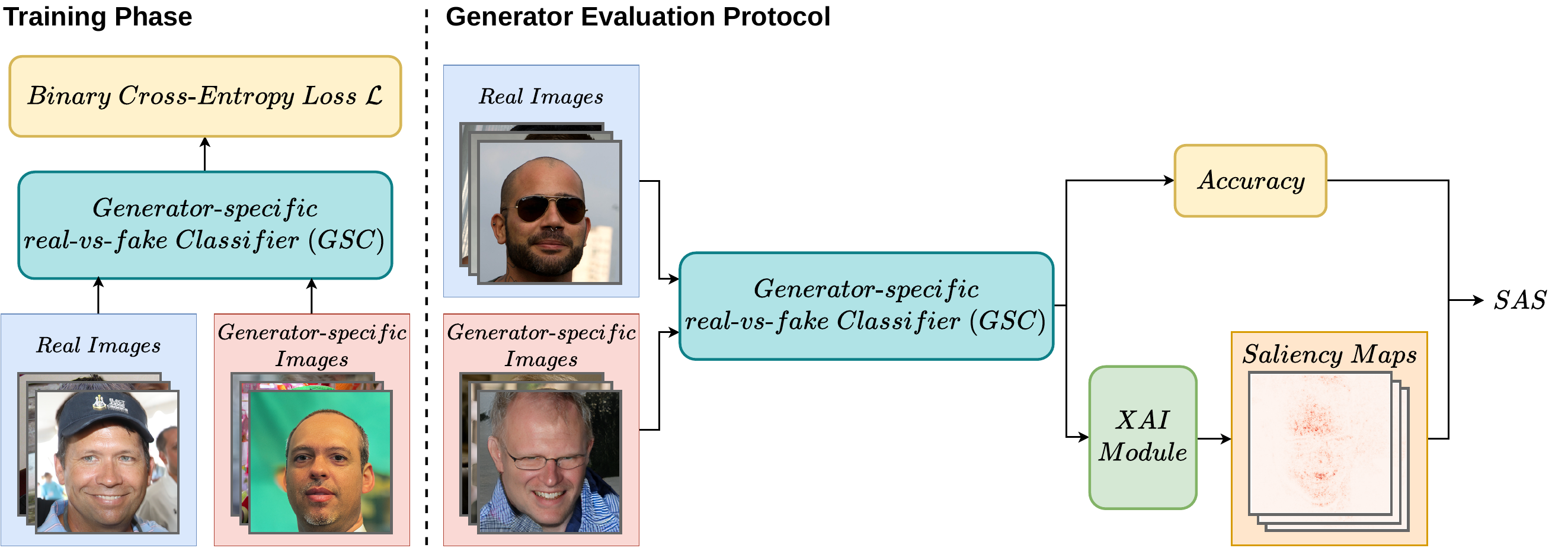}
    \caption{Framework overview. We train a generator-specific real-vs-fake
    classifier (GSC) to distinguish real images from those generated by a
    specific model; then, we combine performance on such task with XAI
    saliency maps to evaluate generator quality.}
    \label{fig:framework}
\end{figure}

Among the most widely used metrics are Fréchet inception distance
(FID)~\cite{DBLP:conf/nips/HeuselRUNH17} and related
variants~\cite{DBLP:conf/cvpr/JayasumanaRVGCK24,
DBLP:conf/nips/SteinCHSRVLCTL23}, as well as complementary measures such as
precision/recall~\cite{DBLP:conf/nips/KynkaanniemiKLL19}
and density/coverage~\cite{DBLP:conf/icml/NaeemOUCY20}. These metrics
typically follow a two-stage design: images are first embedded into a feature
space using a pretrained neural network, and distances between the
distributions of real and generated samples are then computed in that space.
While effective and widely adopted, this design makes evaluation strongly
dependent on the chosen feature extractor. 

With the advent
of self-supervised learning (SSL),
recent protocols often replace
Inception-V3~\cite{DBLP:conf/cvpr/SzegedyVISW16}, trained on
ImageNet-1k~\cite{DBLP:conf/cvpr/DengDSLL009}, with more general-purpose
feature extractors such as
DINOv2~\cite{DBLP:journals/tmlr/OquabDMVSKFHMEA24},
SimCLRv2~\cite{DBLP:conf/nips/ChenKSNH20}, and
CLIP~\cite{DBLP:conf/icml/RadfordKHRGASAM21}. This shift mitigates known
limitations of Inception-based features, as quantified in the benchmark
reported in~\cite{DBLP:conf/nips/SteinCHSRVLCTL23}, which spans multiple
real-image datasets and relies on extensive human evaluations. In that paper,
DINOv2 is reported to be the feature extractor that best correlates with
human ability to distinguish real from generated images for almost all the
aforementioned metrics. At the same time, these results highlight a key
limitation: evaluation performance strongly depends on the chosen
feature extractor. For instance, using SimCLRv2 features shows almost no
correlation with human judgments for all metrics.

Alongside replacing Inception-V3 with large unsupervised or self-supervised models, recent work has questioned FID itself, arguing that its underlying multivariate Gaussian assumption is generally invalid for neural feature representations. Jayasumana et al.~\cite{DBLP:conf/cvpr/JayasumanaRVGCK24} address this issue with CMMD, replacing the Fréchet distance with non-parametric maximum mean discrepancy (MMD) and using CLIP rather than DINOv2 for feature extraction. Thus, the choice of feature extractor remains critical, and metric-human correlation strongly depends on it, as we show in the remainder of this paper.

A further potential limitation of these evaluation protocols is that
representations learned for classification, SSL, or multimodal alignment are
assumed to be suitable for measuring generative realism. Yet, these
objectives differ fundamentally from real-versus-fake
discrimination~\cite{DBLP:conf/iclr/KynkaanniemiKAA23}. Classification and
SSL models are typically trained to be invariant to low-level perturbations
(subtle texture inconsistencies, high-frequency artifacts, or minor
geometric distortions) so long as semantic identity is preserved.
Yet, generative artifacts often appear exactly as such low-level
irregularities, meaning that semantically oriented feature spaces may
suppress the most informative cues for detecting fakes.

As the goal of generative modeling is often perceptual, i.e.,
producing images that appear realistic to humans, a key question is how well
automatic metrics align with human judgments. To quantify realism directly at
the perceptual level, the HYPE benchmark was introduced
in~\cite{DBLP:conf/nips/ZhouGKNFB19}. In particular, the
HYPE$_{\infty}$ protocol models realism assessment as a two-alternative
forced-choice real-vs-fake task: at each trial, a participant is shown either
a real image sampled from the training distribution or a generated sample from
a specific model, and must decide whether the image is real or fake. Models
are then scored by human error rate (HER), defined as the fraction of
incorrect judgments. Large-scale benchmarks such as
\cite{DBLP:conf/nips/SteinCHSRVLCTL23} adopt HYPE$_{\infty}$ to evaluate the
human-alignment of automatic metrics across datasets and model families,
making HER a practical reference signal to assess perceptual realism.

Motivated by the above-mentioned limitations, in this paper we investigate
the viability of novel evaluation strategies that do not rely on external
feature extractors but remain aligned to human judgment. Borrowing ideas from
HYPE$_{\infty}$, where participants are shown an image and must decide
whether it is real or fake, we train a generator-specific classifier (GSC) to
discriminate real images from those produced by a given model. We then
leverage explainable AI (XAI) techniques to analyze the classifier's behavior
and use this analysis for model evaluation. The key intuition is that humans
often rely on the presence of visual artifacts that rarely occur in the real
world, such as poorly defined contours, unnatural textures, or repetitive
patterns~\cite{DBLP:journals/jasis/Goh24}. By explicitly identifying and
characterizing these cues via explainability methods, our framework provides
a general evaluation strategy that more closely mirrors human perceptual
judgments.
To this end, we rely on saliency-based attribution methods, which provide
tools to localize and quantify the evidence used by a model. In particular,
Integrated Gradients (IG)~\cite{DBLP:conf/icml/SundararajanTY17} is well
suited to our setting, as it attributes a model's prediction to input pixels
by integrating gradients along a path from a baseline image to the actual
input. More broadly, saliency and explanation methods such as 
SmoothGrad~\cite{DBLP:journals/corr/SmilkovTKVW17},
Grad-CAM~\cite{DBLP:conf/iccv/SelvarajuCDVPB17}, LayerCAM
\cite{DBLP:journals/tip/JiangZHCW21}, XGrad-CAM
\cite{DBLP:conf/bmvc/FuHDGGL20}, and
SHAP~\cite{DBLP:conf/nips/LundbergL17} have been widely used to inspect the
evidence behind neural-network decisions. In our case, however, saliency maps
are not used merely for qualitative visualization: they become part of the
evaluation process, as they are used to quantify how strongly and where a
GSC detects artifacts in generated images.

Our main contributions are summarized as follows:
\begin{enumerate*}[label=(\roman*), itemjoin={;$\;$ }]
    \item we show that GSCs learn model-specific artifacts and that GSC performance and saliency maps are highly significant for spotting and evaluating these artifacts
    \item we propose an evaluation framework that exploits these observed properties to avoid external pretrained feature extractors
    \item we demonstrate that combining saliency and accuracy leads to a score that is strongly aligned with human perceptual judgments across multiple datasets and generator families
\end{enumerate*}.
We release the source code of our evaluation framework at
\href{https://github.com/EleonoraRistori/sas-eval/}{github.com/EleonoraRistori/sas-eval}.

\section{Methodology}\label{sec:methodology}

Our framework to evaluate a generative model $g$ is illustrated in Figure~\ref{fig:framework}. It consists of two stages: (i) training a GSC to distinguish real images from those generated by $g$; (ii) analyzing the performance of this classifier, as well as the features it uses to make its predictions.

\paragraph{Training Phase}
In the first stage, we train a GSC to solve a binary discrimination problem between real and generated samples. Our classifier receives as input a dataset $D_R$ of real images and a dataset $D_F^{(g)}$ of samples produced by the generative model $g$ that we want to evaluate, and it is trained to distinguish these two datasets by minimizing the binary cross-entropy loss.
Our setup differs fundamentally and deliberately from that of traditional fake-detection systems. In traditional fake-detection, the objective is to train a universal detector capable of identifying generated content originating from an unknown and potentially heterogeneous set of generative models, as the fake distribution is often modeled as $\bigcup\limits_{g'\in\mathcal{G}}D_F^{(g')}$ where $\mathcal{G}$ may be partially unknown and evolve over time.
In contrast, our aim is to evaluate the quality of a single generative model $g$ and therefore, the GCS is by design trained to distinguish between real images and fake images from $g$.
By avoiding exposure to multiple generative sources, the GSC does not learn generic “fakeness” cues that promote broad generalization. Rather, as we will show in the next sections, it captures the distinctive statistical regularities and artifacts that uniquely characterize the outputs of model $g$, which can be further quantified through both performance metrics and interpretability analysis.

\paragraph{Generator Evaluation Protocol}
Once the GSC is trained, we analyze its behavior to extract complementary information for evaluating generator $g$. We consider two main components: (i) the GSC accuracy and (ii) the statistical properties of the saliency maps. Intuitively, if a generator produces highly realistic images, the discrimination task becomes more difficult and accuracy decreases, as happens for human observers. However, neural networks are typically much stronger discriminators than humans and often achieve high accuracy even when perceptual differences are subtle. Therefore, accuracy alone is insufficient to reliably assess generative realism.
To address this limitation, we analyze the statistical properties of saliency maps, which highlight the features the GSC uses to distinguish real images from generated ones. To generate saliency maps, we employ IG~\cite{DBLP:conf/icml/SundararajanTY17}, an explainable AI method that attributes the model’s prediction to its input pixels by integrating gradients along a path from a baseline input to the actual image.
Formally, given a GSC $F$, an input image  $x$, and a baseline $x'$, the integrated gradient along pixel $i,j$ and channel $c$ is defined as:
\begin{equation}
    IG_{i,j,c}(x) \doteq \left(x_{i,j,c} - x'_{i,j,c}\right)  \int_0^1 \frac{\partial F (x' + (x-x')\sigma)}{\partial x_{i,j,c}} d\sigma 
\end{equation}
This yields a saliency map that highlights the regions of the input most responsible for the model’s decision, which we can use to identify characteristic artifacts of each generative model and effectively evaluate the quality of its generations.
We also evaluate additional saliency techniques in 
Section~\ref{sec:xai_analysis},
demonstrating that the proposed framework remains stable and consistent across different saliency map methods, thereby confirming its robustness and general applicability.
The saliency component for $g$, given a set $\mathcal{X}_g$ of real and fake images (generated by $g$), is finally obtained by aggregating IG values:
\begin{equation}
    s(g) = \frac{w}{|\mathcal{X}_g|}\sum_{x\in\mathcal{X}_g} \median_{i,j} \left( \sum_c \left|IG_{i,j,c}(x)\right|\right)
\label{eq:s(g)}
\end{equation}
where $w$ is a weight to rescale $s(g)$ in a predefined range, which only depends on the backbone used.
We experiment with a saliency-accuracy score (SAS) that combines classification accuracy $a(g) \in [0,1]$ calculated on the same set $\mathcal{X}_g$ with the saliency component $s(g)$, defined as:
\begin{equation}
    SAS(g) = a(g) + a(g)^\alpha(1-s(g)),
\end{equation}
where $\alpha \gg 1$ is a parameter set to trade off saliency with respect to accuracy. SAS is designed to anticorrelate with the human error rate. Intuitively, when accuracy is low the discrimination task is difficult, suggesting that the generated samples are realistic. At the same time, the contribution of saliency maps is down-weighted through the $a^\alpha$ term, since a lower performance is likely to be associated with less informative features.

\section{Experimental Findings}
\subsection{Experimental Setup}
\label{sec:experimental_setup}

Following the benchmark introduced in \cite{DBLP:conf/nips/SteinCHSRVLCTL23}, we conduct experiments on four natural image datasets: CIFAR10 \cite{krizhevsky2009learning}, FFHQ \cite{DBLP:conf/cvpr/KarrasLA19}, ImageNet \cite{DBLP:conf/cvpr/DengDSLL009}, and LSUN Bedroom \cite{DBLP:journals/corr/YuZSSX15}. For each dataset, we consider multiple generative models (see Table~\ref{tab:gen_models}) trained on the corresponding data distribution, covering a broad spectrum of paradigms, including diffusion models, GANs, VAEs, normalizing flows, and Transformer-based architectures.

% \paragraph{Datasets and generators}
% \label{sec:dataset_details}

% Table~\ref{tab:gen_models} lists the generative models considered for each of the four datasets in our evaluation experiments. 

\begin{table}
    \centering
    \setlength{\tabcolsep}{5pt}
    \caption{Generative models considered for each dataset. For every model listed, a dedicated classifier is trained to distinguish between real images from the corresponding dataset and images generated by that specific model.}
    \begin{adjustbox}{max width=0.92\textwidth,center}
    \begin{tabular}{llll} 

    \toprule
    \textbf{CIFAR10} & \textbf{FFHQ} & \textbf{ImageNet} & \textbf{LSUN Bedroom }\\
    \midrule
    ACGAN \cite{DBLP:conf/nips/KangSCP21}        & Efficient-vdVAE \cite{DBLP:journals/corr/abs-2203-13751}   & ADM \cite{DBLP:conf/nips/DhariwalN21}              & ADM-dropout \cite{DBLP:conf/nips/DhariwalN21} \\
    BigGAN  \cite{DBLP:conf/iclr/BrockDS19}     & InsGen  \cite{DBLP:conf/nips/YangSXZ21}          & ADMG \cite{DBLP:conf/nips/DhariwalN21}             & DDPM \cite{DBLP:conf/nips/HoJA20}\\
    iDDPM-DDIM \cite{DBLP:conf/icml/NicholD21}  & LDM \cite{DBLP:conf/cvpr/RombachBLEO22}              & ADMG-ADMU  \cite{DBLP:conf/nips/DhariwalN21}       & Diff-Proj-GAN \cite{DBLP:conf/iclr/WangZHCZ23}\\
    LOGAN \cite{DBLP:journals/corr/abs-1912-00953}       & Projected-GAN \cite{DBLP:conf/iccv/PeeblesX23}    & BigGAN \cite{DBLP:conf/iclr/BrockDS19}           & iDDPM \cite{DBLP:conf/icml/NicholD21}\\
    LSGM-ODE \cite{DBLP:conf/nips/VahdatKK21}    & StyleGAN-XL \cite{DBLP:conf/siggraph/SauerS022}      & DiT-XL-2  \cite{DBLP:conf/iccv/PeeblesX23}        & Projected-GAN \cite{DBLP:conf/iccv/PeeblesX23}\\
    MHGAN  \cite{DBLP:conf/icml/TurnerHFSY19}      & StyleGAN2-ada \cite{DBLP:conf/nips/KarrasAHLLA20}     & DiT-XL-2-guided \cite{DBLP:conf/iccv/PeeblesX23}  & StyleGAN \cite{DBLP:conf/cvpr/KarrasLA19}\\
    NVAE  \cite{DBLP:conf/nips/VahdatK20}       & StyleNAT  \cite{DBLP:journals/corr/abs-2211-05770}        & GigaGAN  \cite{DBLP:conf/cvpr/KangZ0PSPP23}          & Unleash-Trans \cite{DBLP:conf/eccv/Bond-TaylorH0BW22}\\
    PFGM++ \cite{DBLP:conf/icml/XuLTTTJ23}      & StyleSwin \cite{DBLP:conf/cvpr/ZhangGZBCWWG22}        & LDM \cite{DBLP:conf/cvpr/RombachBLEO22}              &   \\
    ReACGAN  \cite{DBLP:conf/nips/KangSCP21}    & Unleash-Trans \cite{DBLP:conf/eccv/Bond-TaylorH0BW22}     & Mask-GIT   \cite{DBLP:conf/cvpr/ChangZJLF22}       &   \\
    RESFLOW \cite{DBLP:conf/nips/ChenBDJ19}      &                   & RQ-Transformer \cite{DBLP:conf/cvpr/LeeKKCH22}   &   \\
    StyleGAN-XL \cite{DBLP:conf/siggraph/SauerS022} &                   & StyleGAN-XL  \cite{DBLP:conf/siggraph/SauerS022}     &   \\
    StyleGAN2-ada \cite{DBLP:conf/nips/KarrasAHLLA20}&                   &                   &   \\
    WGAN-GP \cite{DBLP:journals/corr/GulrajaniAADC17}     &                   &                   &   \\
    \bottomrule
    \end{tabular}
    \end{adjustbox}
    \label{tab:gen_models}
\end{table}

\paragraph{Implementation details}
\label{sec:implementation_details}
For each generative model, we train its corresponding GCS for 30 epochs on a balanced dataset consisting of 45,000 real images sampled from the real dataset $D_R$ and 45,000 generated images from the model $g$ under evaluation $D_F^{(g)}$. An additional 5,000 real and 5,000 fake images per model were held out as a validation set and used to select the best-performing model across the 30 training epochs, allowing us to monitor validation performance and mitigate overfitting. 
We remark that these data requirements are comparable to those of FID in terms of the number of images needed for reliable estimation. 

For the three high-resolution datasets, the GSC backbone is a ResNet18, whereas for CIFAR10 we adopt a smaller ResNet9, which is designed for $32\times32$ inputs.
To account for these architectural differences, the saliency component $s(g)$ is rescaled to compensate for differences in magnitude induced by the two architectures. For ResNet18, $s(g)$ is multiplied by a factor of $w=10$, whereas for the ResNet9 used for CIFAR10, the scaling factor is set to $0.1$. This normalization ensures a balanced contribution of the saliency term with respect to the accuracy one in SAS.
In both cases, $\alpha$ is set to $100$, although SAS is not particularly sensitive to the $\alpha$ value (see~\ref{app:varying_alpha}).
We ran all our experiments using a single RTX 5000 Ada with 32 GB of VRAM, where evaluating a single model takes around $\sim35$ minutes, which is not much more expensive than the $\sim25$ minutes necessary for calculating FD on DINOv2 features.

\subsection{\textbf{SAS Is Aligned with Human Judgment}}
\label{sec:combining_acc_sal}

\begin{figure}[htbp]
    \centering
    \includegraphics[width=0.9\linewidth]{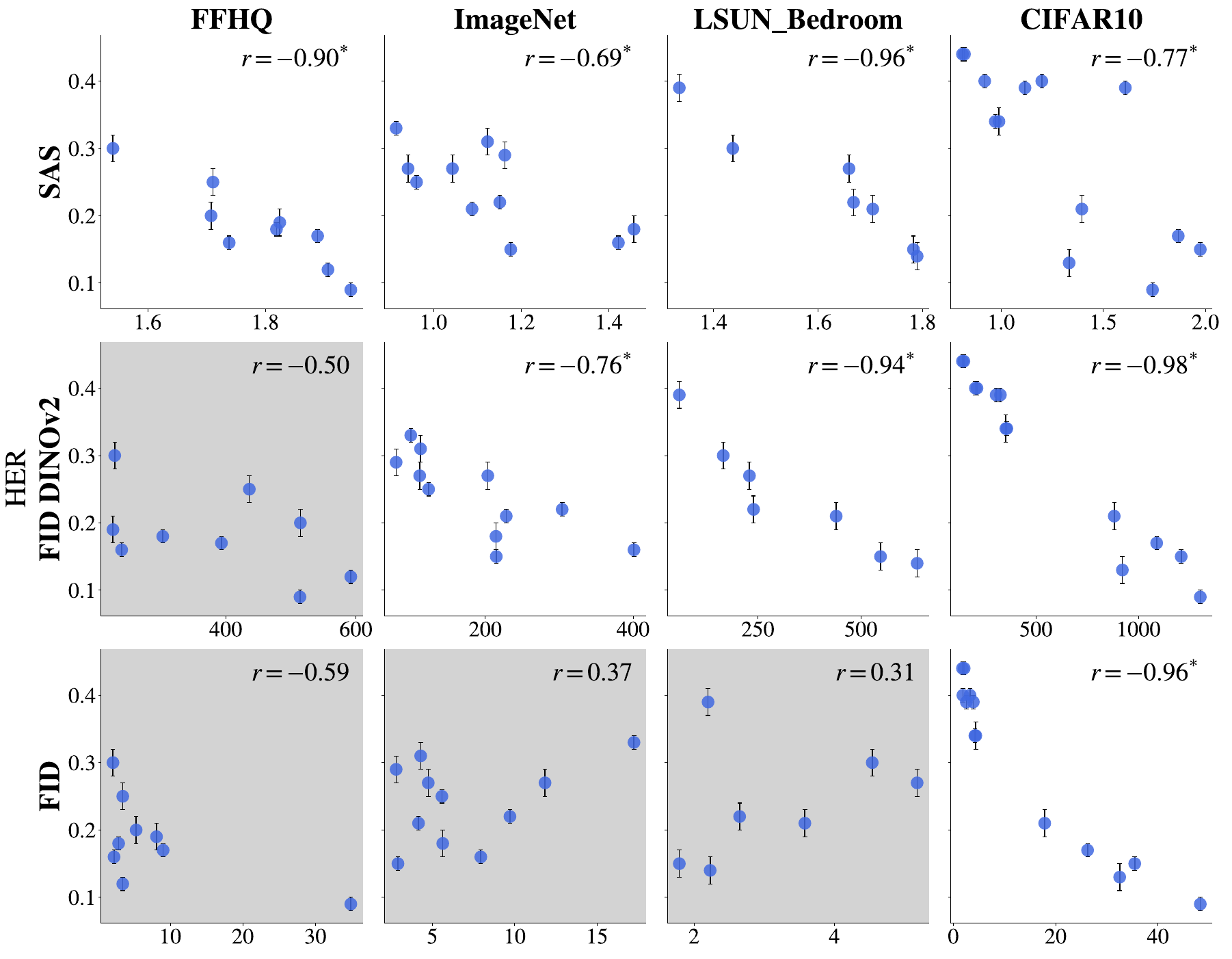}
    \caption{Metrics vs HER. Both FID and SAS are negatively correlated with HER. $r$ denotes the Pearson correlation, shaded background indicates correlations that are not statistically significant ($p \leq 0.05$) with respect to HER. The standard deviation is calculated across different sets of images.}
    \label{fig:correlation_results_ours}
\end{figure}

Figure~\ref{fig:correlation_results_ours} reports the relationship between HER and SAS in comparison with (i) the traditional FID score and (ii) the Fréchet Distance  introduced in~\cite{DBLP:conf/nips/SteinCHSRVLCTL23}, which leverages DINOv2 as backbone. Across all datasets, SAS consistently exhibits a strong anti-correlation with human evaluations and is the only one to achieve statistical significance with HER for every dataset. In contrast, competing metrics exhibit weaker and, in several cases, non-significant correlations.

\paragraph{Comparison with additional metrics}
\label{sec:other_metrics}
Table~\ref{tab:other_metrics} reports the Pearson correlation between the human error rate and several state-of-the-art metrics. Our SAS score achieves a consistently strong and statistically significant correlation, whereas many competing approaches exhibit clear dependence on the choice of pretrained feature extractor, leading to less stable and less reliable performance across datasets.

\begin{table}[t]
\centering
\setlength{\tabcolsep}{2pt}
\caption{Pearson correlation ($r$) between different metrics and the human error rate.}
\begin{adjustbox}{max width=\textwidth,center}
\begin{tabular}{lcccccccccccc}
\toprule
& \multicolumn{3}{c}{\textbf{CIFAR10}} & \multicolumn{3}{c}{\textbf{ImageNet256}} & \multicolumn{3}{c}{\textbf{LSUN Bedroom}} & \multicolumn{3}{c}{\textbf{FFHQ256}} \\
\cmidrule(lr){2-4} \cmidrule(lr){5-7} \cmidrule(lr){8-10} \cmidrule(lr){11-13}
\textbf{Metric} & DINOv2 & SimCLRv2 & Inception-V3 & DINOv2 & SimCLRv2 & Inception-V3 & DINOv2 & SimCLRv2 & Inception-V3 & DINOv2 & SimCLRv2 & Inception-V3 \\
\midrule
FD \cite{DBLP:conf/nips/HeuselRUNH17}            & -0.98 & -0.37 & -0.96 & -0.76 & $\;$0.52  & $\;$0.37 & -0.93 & -0.54 & -0.15 & -0.50 & $\;$0.19 &-0.61 \\
FD$_{\infty}$ \cite{DBLP:conf/cvpr/ChongF20} & -0.98 & -0.37 & -0.96 & -0.77 & $\;$0.52 & $\;$0.37 & -0.93 & -0.54 & -0.15 & -0.50 & $\;$0.20 & -0.61 \\
KD  \cite{DBLP:conf/iclr/BinkowskiSAG18}          & -0.96 & -0.38 & -0.95 & -0.78 & $\;$0.47 & $\;$0.47 & -0.92 & -0.28 & -0.02 & -0.47 & $\;$0.30  & -0.60 \\
FLS \cite{DBLP:conf/nips/JiralerspongBGQ23}          &  $\;$0.98 & -0.76 & $\;$0.97 &  $\;$0.84 & -0.45 &  $\;$0.01 &   --   &   --  &   --  &   --  &   --  &   -- \\
Precision \cite{DBLP:conf/nips/KynkaanniemiKLL19}    & -0.71 & -0.85 & $\;$0.86 &  $\;$0.74 & -0.22  & $\;$0.17 &  $\;$0.49  & -0.44 &  $\;$0.16 & -0.08 & $\;$0.38 &  -0.27 \\
Recall \cite{DBLP:conf/nips/KynkaanniemiKLL19}       &  $\;$0.94 & $\;$0.66 & $\;$0.81 &  $\;$0.34  & -0.12 & -0.15  &  $\;$0.98 & $\;$0.58 & $\;$0.21  &  $\;$0.34 & -0.16 & $\;$0.59 \\
Density \cite{DBLP:conf/icml/NaeemOUCY20}      & -0.86 & -0.82 &   $\;$0.94 &  $\;$0.58  & -0.19 & -0.12 &  $\;$0.48  & -0.48 &   $\;$0.05 & -0.15 & $\;$0.45 &  -0.33 \\
Coverage \cite{DBLP:conf/icml/NaeemOUCY20}      &  $\;$0.97 & $\;$0.36 &  $\;$0.98 &  $\;$0.81 & -0.51 & -0.13 &  $\;$0.92 & $\;$0.60 &  $\;$0.22 &  $\;$0.41 & $\;$0.29 &  $\;$0.63 \\
AuthPct \cite{DBLP:conf/icml/AlaaBSS22}      &  $\;$0.82 & $\;$0.66 & -0.85 & -0.63 & $\;$0.27  & -0.42 & -0.78 & $\;$0.39 & -0.03 &  $\;$0.05 & -0.54 & -0.18 \\
C$_T$  \cite{DBLP:journals/corr/abs-2004-05675}       & -0.44 & $\;$0.55  & -0.98  & -0.79 & $\;$0.38 & -0.24 &   --   &   --  &   --  &   --  &   --  &   --  \\
C$_T$ mod. \cite{DBLP:conf/nips/SteinCHSRVLCTL23}    & -0.83 & $\;$0.12 & -0.52 & -0.65 & -0.49 & -0.35 &   --   &   --  &   --  &   --  &   --  &   -- \\
FLS-POG \cite{DBLP:conf/nips/JiralerspongBGQ23}      & -0.75 & $\;$0.10 &  $\;$0.09 &  $\;$0.40  & $\;$0.39 &  $\;$0.35 &   --   &   --  &   --  &   --  &   --  &   -- \\
\midrule

CMMD \cite{DBLP:conf/cvpr/JayasumanaRVGCK24}
& \multicolumn{3}{c}{-0.96}
& \multicolumn{3}{c}{-0.67}
& \multicolumn{3}{c}{-0.98}
& \multicolumn{3}{c}{-0.20} \\
SAS (Ours)
& \multicolumn{3}{c}{-0.77}
& \multicolumn{3}{c}{-0.69}
& \multicolumn{3}{c}{-0.96}
& \multicolumn{3}{c}{-0.90} \\
\bottomrule
\end{tabular}
\end{adjustbox}
\label{tab:other_metrics}
\end{table}

These results suggest that combining classification performance with interpretability-based information derived from saliency maps provides a meaningful and complementary perspective for assessing generative model realism.
In the following sections, we validate the individual components of SAS by analyzing the behavior of our GSCs and by motivating the choice of classification performance and saliency as key elements to evaluate generative models.

\subsection{The Relevance of the OOD Performance}
\label{sec:performance_analysis}

\begin{figure}[htbp]
    \centering
    \includegraphics[width=0.97\linewidth]{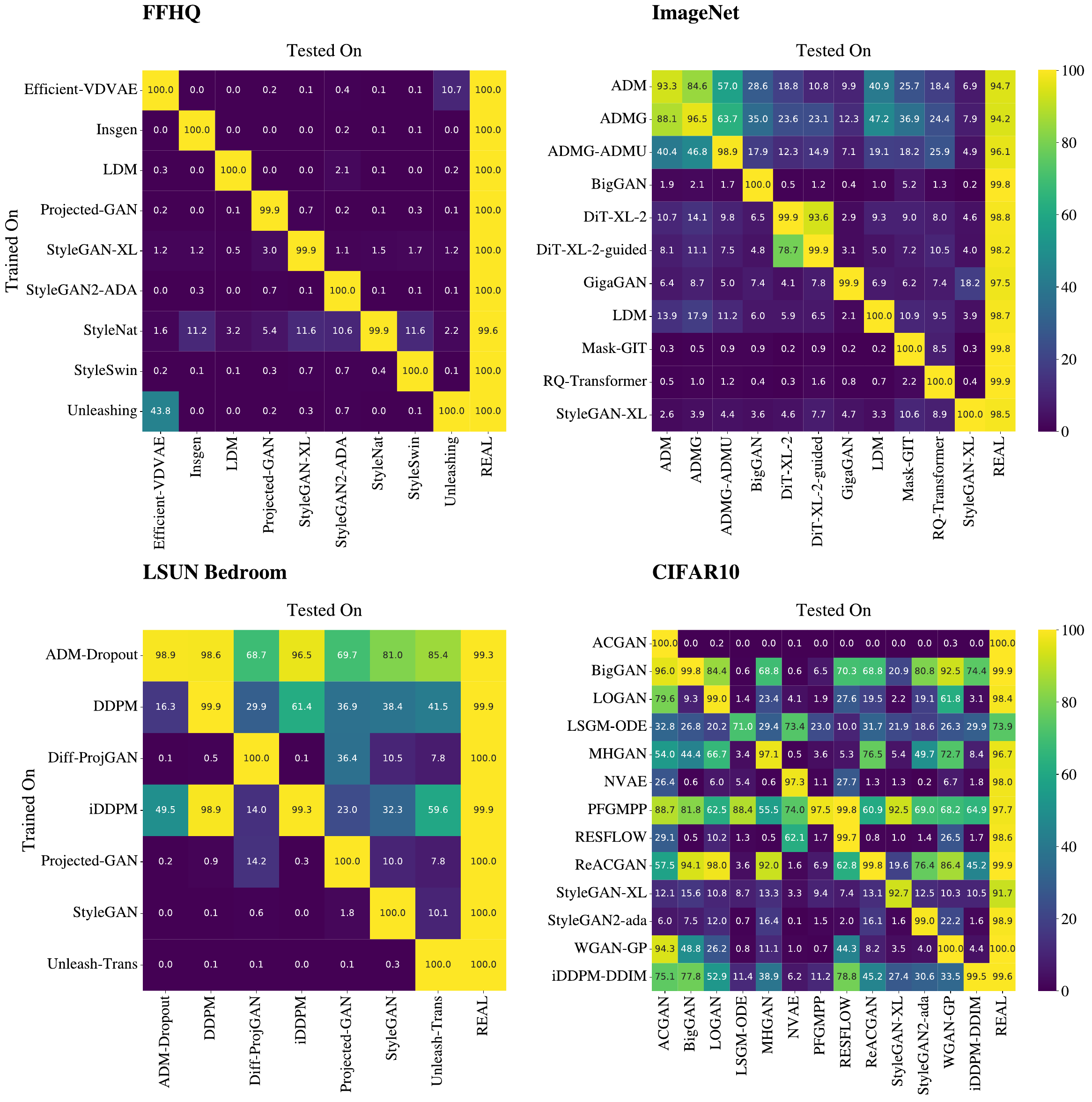}
    \caption{Cross/self generator recalls (CSGR). Entry $(i,j)$ shows the recall of the GSC trained on generator $i$ and tested on images from generator $j$ (fake being the positive class). The last column reports the recall of each GSC when tested on real images.}
    \label{fig:confusion_matrix}
\end{figure}

In this section, we analyze the performance of our GSC, the first component of SAS. To this end, we report the \emph{cross/self-generator recalls} (CSGR) in Figure~\ref{fig:confusion_matrix}, using a pool of $M$ generators. Specifically, the value in the $i$-th row and $j$-th column represents the cross-generator (self-generator when $i=j$) recall of the GSC trained on images from generator $i$ and tested on images from generator $j$ (with $i,j \in {1,\dots,M}$), considering fake images as the positive class. The $(M+1)$-th column shows the recall of each GSC when evaluated on real images. In all experiments, recalls are computed over the 5,000 test images.
A clear pattern emerges: self-generator recalls strongly dominate over cross-generator ones. This is particularly evident for FFHQ, where the CSGR is perfectly diagonal, but the same trend is consistently observed across all datasets. Such a behavior suggests a specific underlying mechanism: each GSC appears to learn a default ``real'' class and to rely on generator-specific cues to label an image as fake. If this were not the case, performance on samples generated by unseen generators would be close to random (i.e., around $50\%$ recall), or even higher if the GSC had learned a more general notion of ``realism'' based on real data, which is the behaviour shown by GSCs trained to distiguish real from fake features in the DINOv2 embedding space as shown in~\ref{app:DINOv2 GSCs}. Instead, here cross-generator recalls are often close to zero.

As an additional observation, despite the general tendency of the CSGRs to assume a diagonal structure, some rows exhibit higher cross-generator recall. This fact suggests that, in certain cases, a classifier trained to distinguish real images from the fake outputs of a single generative model is more robust to out-of-distribution samples than others.
Since the task is always the same—distinguishing real from fake for a specific generator—and the training procedure for each GSC is identical, these differences in performance can be attributed solely to the nature of the generated samples.

\paragraph{Cross-dataset evaluation}
\label{sec:cross-dataset}
A similar pattern of limited generalization holds in cross-dataset evaluations.  Table~\ref{tab:cross-dataset_accuracy} reports the cross-dataset recall, computed by evaluating our GSCs on generated images from datasets different from those used during training. The results indicate that the GSCs exhibit very limited generalization, as recall on out-of-training generated samples remains consistently low for every dataset, while the one on real samples is consistently high. This behavior supports our hypothesis that the GSCs primarily capture generator-specific artifacts present in the training distributions, while assigning a default “real” label to samples that do not match these learned signatures. 
\begin{table}[t]
\centering
\renewcommand{\arraystretch}{1.05}
\setlength{\tabcolsep}{2.5pt}
\caption{Cross-dataset recall. GSCs are evaluated on generated images from other datasets. \textit{Fake in} denotes the recall on fake samples generated by the same model used during training, whereas \textit{Fake out} refers to the recall on fake samples produced by generative models not seen during training.}
\begin{tabular}{llcccc}
\toprule
\multirow{2}{*}{\textbf{Training}} 
& \multirow{2}{*}{\textbf{Recall}} 
& \multicolumn{4}{c}{\textbf{Test dataset}} \\
\cmidrule(lr){3-6}
& & \textbf{LSUN} & \textbf{ImageNet} & \textbf{FFHQ} & \textbf{CIFAR10} \\
\midrule
\multirow{3}{*}{LSUN}
 &  Real                  & 98.36     & 92.37     & 87.81     & 95.63 \\
 &  Fake out              & 22.85     & 13.05     & 16.82     & 5.42 \\
 &  Fake in               & 92.96     & -         & -         & - \\
\addlinespace[2pt]
\multirow{3}{*}{ImageNet} 
 &  Real                  & 94.60     & 95.33     & 93.75     & 91.41 \\
 &  Fake out              & 16.90     & 14.42     & 12.29     & 8.88 \\
 &  Fake in               & -         & 91.07     & -         & - \\
\addlinespace[2pt]
\multirow{3}{*}{FFHQ} 
 &  Real                  & 83.02     & 90.30     & 99.32     & 72.09 \\
 &  Fake out              & 20.67     & 12.38     & 0.03      & 30.31 \\
 &  Fake in               & -         & -         & 99.49     & - \\
\addlinespace[2pt]
\multirow{3}{*}{CIFAR10} 
 &  Real                  & 85.98     & 83.89     & 83.59     & 94.45 \\
 &  Fake out              & 17.58      & 18.61     & 17.70     & 31.03 \\
 &  Fake in               & -         & -         & -         & 96.53 \\
 
\bottomrule
\end{tabular}

\label{tab:cross-dataset_accuracy}
\end{table}
Previous works have observed a similar phenomenon in substantially different settings:~\cite{NEURIPS2024_6dddcff5} in the domain of natural images and~\cite{10890655} in the context of audio generation. In both cases, classifiers achieve nearly $100\%$ recall on real samples and close-to-zero recall on fake samples outside the training distribution. In practice, this means that any input not belonging to the specific fake distribution seen during training is labeled as real.
Taken together, these results indicate that the classifier does not learn a general notion of ``fakeness'' or ``realism''. Rather, it detects generator-specific artifacts. When such artifacts are absent, as in the case of unseen generators or different datasets, the GSC defaults to predicting the real class.

\paragraph{Cross-generator recall and HER}
\label{sec:cross-gen-rec}
 Simply examining the cross-generator recall it is possible to correctly identify almost all the models that humans perceive as better, i.e., those on which human subjects make higher errors in real/fake attribution, as shown in Figure~\ref{fig:quadranti}. For each dataset, the cross-generator recall is plotted against HER, revealing that GSCs tend to cluster models into two main groups: (i) models that generate high-quality images, where HER is above $25\%$—these are also the models on which classifiers have learned to generalize; and (ii) models that generate images that are easy to distinguish from real ones, where humans perform almost perfectly, and classifier generalization is minimal. 
The two clusters were identified automatically by fitting a Gaussian mixture model (GMM) on the HER and cross-generator recall values.

\subsection{Analysis of XAI Saliency Maps}
\label{sec:xai_analysis}

Human perceptual studies have shown that observers often identify generated
images by detecting low-level visual artifacts, such as unnatural textures,
local inconsistencies, or structural irregularities
\cite{DBLP:journals/jasis/Goh24}. As discussed in the previous section, our
GSCs exhibit a similar sensitivity to generator-specific cues, as demonstrated
by their limited generalization to images produced by unseen generators.

The second component of SAS therefore exploits saliency maps to characterize
the evidence used by each GSC when distinguishing real from generated images.
In particular, Integrated Gradients allows us to quantify the contribution of
individual image regions to the classifier decision. Rather than using these
maps only for qualitative interpretation, we aggregate their attribution
magnitude through the saliency component $s(g)$ defined in
Equation~\ref{eq:s(g)} and investigate whether this quantity carries
information about the perceptual quality of the corresponding generator.

\begin{figure}[t]
    \centering
    \includegraphics[width=\linewidth]{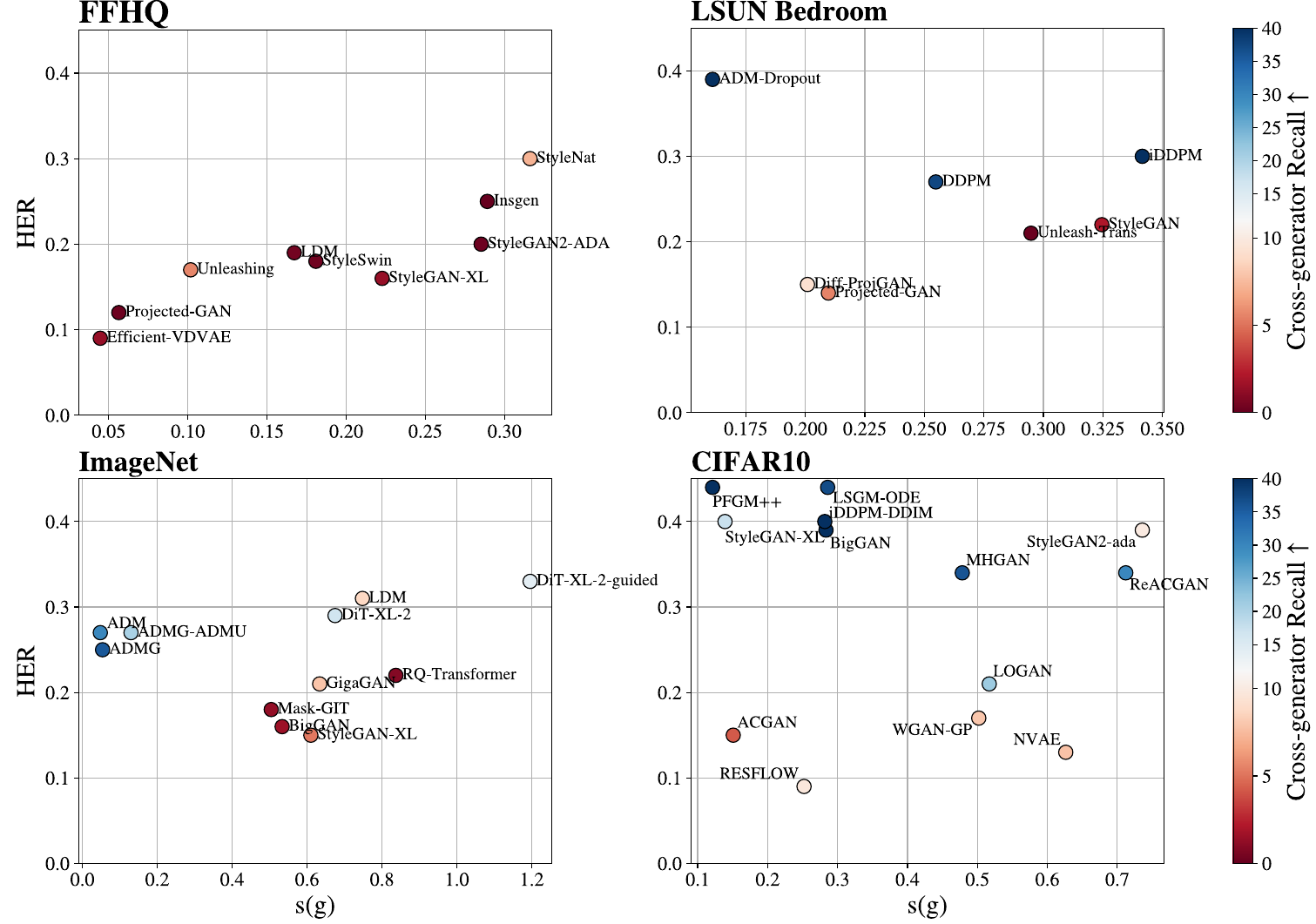}
    \caption{$s(g)$ vs HER. \textcolor{myred}{Red points} denote models for which the GSC achieves limited cross-generator generalization, indicating strong learning of model-specific artifacts. \textcolor{myblue}{Blue points} correspond to GSCs with greater generalization across generators, suggesting that they learn a more generator-agnostic notion of ``fakeness''.}
    \label{fig:corr_saliency_plot}
\end{figure}

Figure~\ref{fig:corr_saliency_plot} reports $s(g)$ against the human error rate
(HER) for all considered generators. To account for the different behaviors
observed in the cross-generator experiments, points are color-coded according
to the cross-generator recall of the corresponding GSC.
\textcolor{myred}{Red points} denote classifiers exhibiting limited
cross-generator generalization, suggesting that their decisions predominantly
rely on artifacts specific to the generator used during training.
\textcolor{myblue}{Blue points}, instead, correspond to classifiers showing
greater generalization across different generators.

For GSCs with limited cross-generator generalization, a clear relationship
emerges between saliency magnitude and HER. Generators that are more difficult
for humans to distinguish from real images tend to induce larger aggregate
attribution values in the corresponding classifier. This suggests that, when
the classifier is effectively specialized to a particular generator,
the strength of the discriminative evidence captured by its saliency maps
contains information complementary to classification accuracy.

Conversely, this relationship is weaker for GSCs exhibiting stronger
cross-generator generalization. In these cases, the classifier appears to rely
on features that are less tightly associated with the artifacts of the
specific generator under evaluation and that transfer more readily to samples
from other generative models. Consequently, the corresponding saliency
statistics are less informative about the quality of that individual
generator.

These observations motivate the joint use of classification performance and
saliency information in SAS. Accuracy quantifies how easily a generator can be
distinguished from real data, while the saliency component provides additional
information about the discriminative evidence exploited by the GSC.
Their combination therefore provides a more informative evaluation signal than
either component considered independently.

\subsection{Robustness to Design Choices}
\label{sec:robustness_design}

\paragraph{Aggregation of saliency values}
\label{sec:varying_aggr_method}
Table~\ref{tab:aggregation_methods} reports the Pearson correlation with the human error rate obtained by SAS when using different aggregation strategies for the saliency maps in the computation of the component $s(g)$. We evaluate a range of aggregation methods, including mean, median, L1 and L2 norms, robust maximum and minimum, sum, and entropy, to assess the robustness of the proposed score. Empirically, the median emerges as the most effective aggregation strategy. However, all considered methods yield consistently high and statistically significant correlations, further supporting the robustness of SAS across different aggregation choices.

\begin{table}[t]
\centering
\setlength{\tabcolsep}{2.5pt}
\caption{Pearson correlation $r$ between SAS and HER by varying the aggregation method for the saliency maps in the computation of $s(g)$ presented in Equation~\ref{eq:s(g)}. Values marked with * are statistically significant.}
\label{tab:results}
\begin{tabular}{lcccc}
\toprule
\textbf{Aggregation Method} & \textbf{CIFAR10} & \textbf{ImageNet} & \textbf{FFHQ} & \textbf{LSUN} \\
\midrule
Median  & -0.77* & -0.69* & -0.90* & -0.96* \\
Mean    & -0.77* & -0.58$\;$       & -0.86* & -0.89* \\
L1 norm     & -0.73* & -0.65* & -0.87* & -0.94* \\
L2 norm    & -0.57* & -0.60* & -0.81* & -0.89* \\
Max     & -0.65* & -0.56$\;$      & -0.78* & -0.87* \\
Min     & -0.78* & -0.63*      & -0.86* & -0.94* \\
Sum     & -0.73* & -0.60* & -0.87* & -0.94* \\
Entropy & -0.78* & -0.45$\;$      & -0.62$\;$       & -0.93* \\
\bottomrule
\end{tabular}
\label{tab:aggregation_methods}
\end{table}

\paragraph{Alternative XAI techniques}
\label{sec:other_XAI_methods}

\begin{figure}[htbp]
    \centering
    \includegraphics[width=0.99\textwidth]{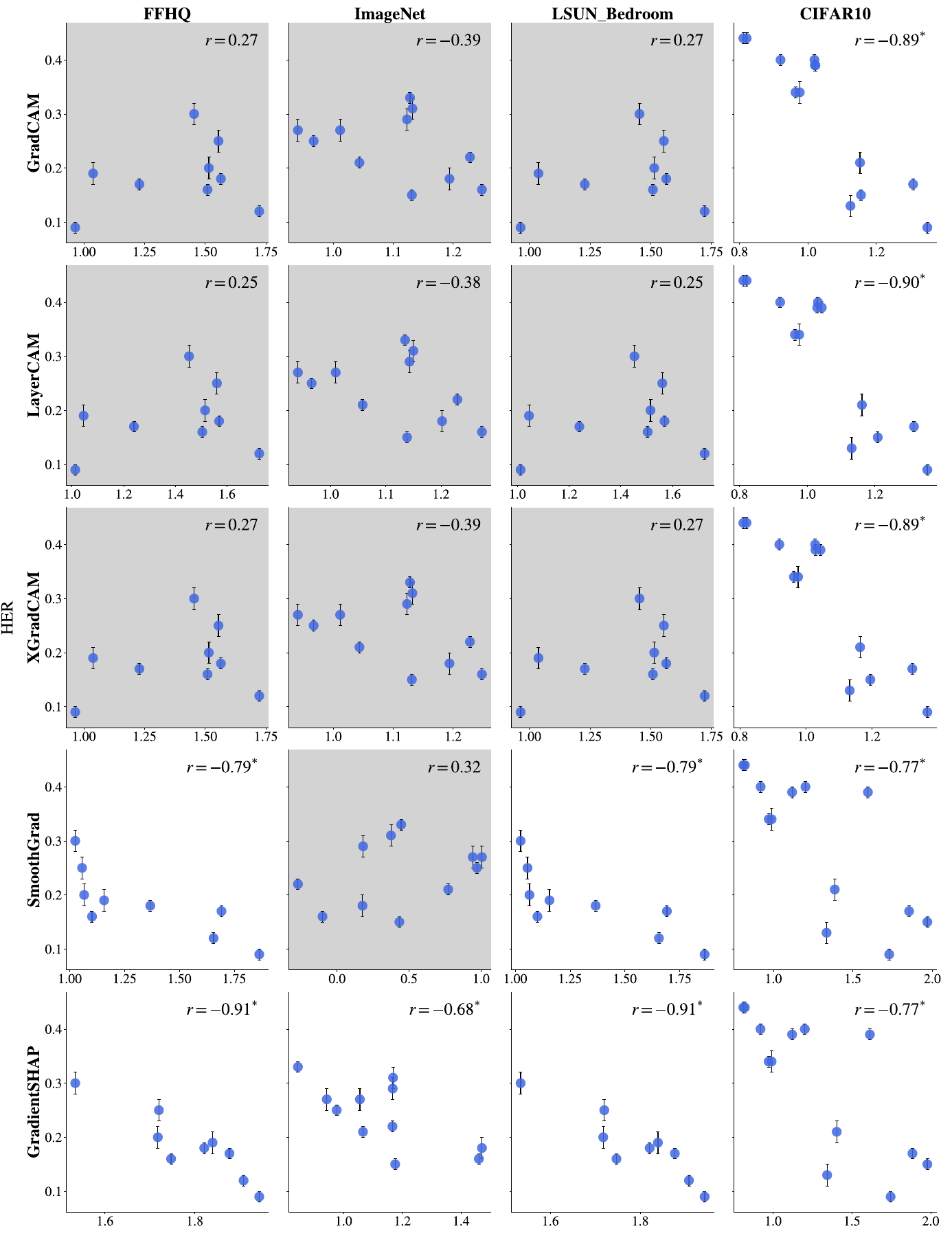}
    \caption{SAS vs HER, where $s(g)$ is calculated using different techniques. The coefficient $r$ denotes the Pearson correlation, and panels with a shaded background indicate correlations that are not statistically significant ($p \leq  0.05$) with respect to HER. GradientSHAP and SmoothGrad maintain strong correlations with HER, while CAM-based methods show weaker results, likely due to their coarse spatial resolution.\label{fig:xai_metrics}}
\end{figure}

To demonstrate that our framework is not tied to a specific attribution method, we replicate the analysis of Section~\ref{sec:combining_acc_sal} using five alternative saliency techniques: GradCAM~\cite{DBLP:conf/iccv/SelvarajuCDVPB17}, LayerCAM~\cite{DBLP:journals/tip/JiangZHCW21}, XGradCAM~\cite{DBLP:conf/bmvc/FuHDGGL20}, GradientSHAP~\cite{DBLP:conf/nips/LundbergL17}, and SmoothGrad~\cite{DBLP:journals/corr/SmilkovTKVW17}.

Figure~\ref{fig:xai_metrics} reports the resulting correlations between SAS and HER.
GradientSHAP and SmoothGrad show strong and significant correlations with human judgments,
while CAM-based methods do not.

\begin{figure}[htbp]
    \centering
    \includegraphics[width=0.8\textwidth]{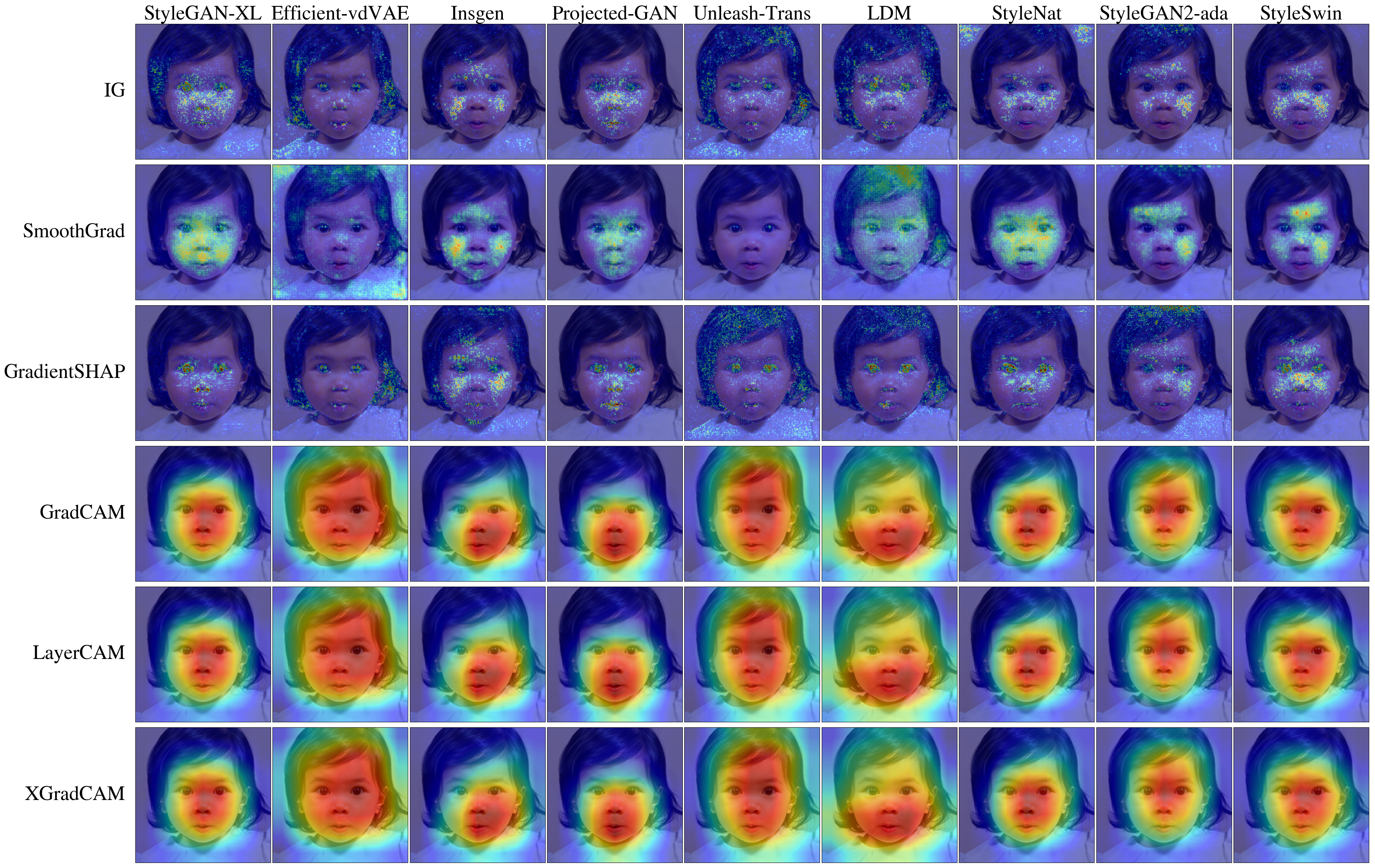}
    \caption{Qualitative comparison of saliency maps produced by different attribution methods. CAM-based approaches generate coarse maps often focusing on similar central regions across generators. Integrated Gradients (IG), SmoothGrad and GradientSHAP instead produce finer and more diverse attribution patterns, which better capture generator-specific artifacts.\label{fig:saliency_comparison}}
\end{figure}

Figure~\ref{fig:saliency_comparison} shows a qualitative comparison of saliency maps across methods. CAM-based techniques produce notably coarser maps, as they are computed at a low spatial resolution (typically $7\times7$) and tend to highlight similar central regions regardless of the generator $g$ on which the GSC is trained, resulting in poorly discriminative patterns across generators and limiting their ability to capture generator-specific artifacts.
In contrast, SmoothGrad and GradientSHAP produce finer-grained and more diverse attribution patterns that better capture generator-specific artifacts, particularly in the high-resolution dataset. For CIFAR-10, however, where the images are only $32\times32$ pixels, the lower-resolution attribution maps produced by CAM-based methods appear to be quite effective.

\subsection{Alternative Aggregation via Cross-Generator Recall}
\label{sec:other_aggr_method}
\begin{figure}[htbp]
    \centering
    \includegraphics[width=0.6\columnwidth]{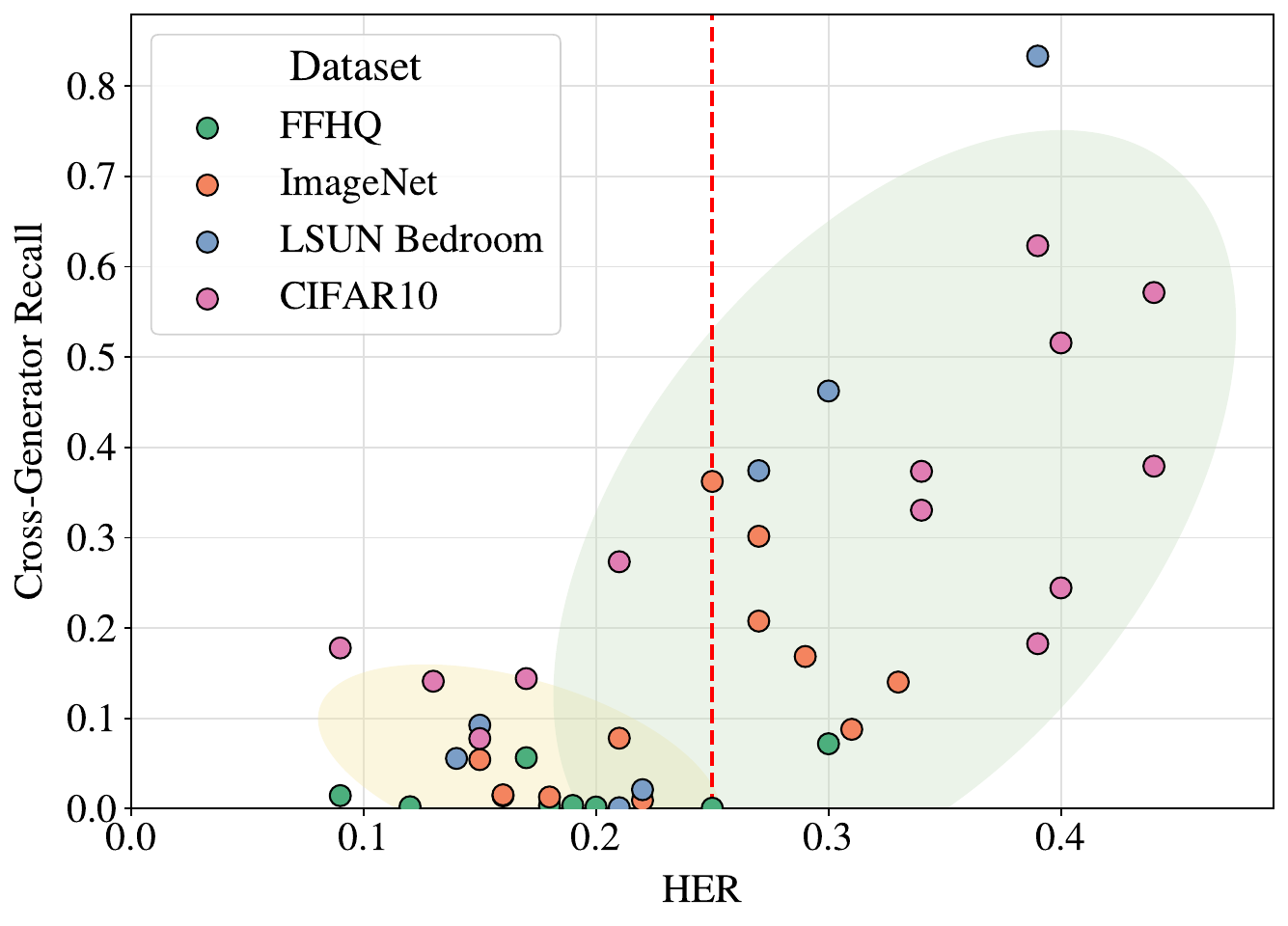}
    \caption{Cross-generator recall vs HER. GSCs separate models into two groups: high-quality generators, where human error exceeds $25\%$ and GSCs show stronger generalization; and low-quality generators, where human accuracy is nearly perfect, and GSC generalization is minimal.\label{fig:quadranti}}
\end{figure}

Since cross-generator recall emerged as a significant factor in determining the perceived realism of generated images (see Section~\ref{sec:performance_analysis} and Figure~\ref{fig:quadranti}), we introduce a formulation of the evaluation score that explicitly uses this signal. Rather than a fully alternative aggregation strategy, this approach should be viewed as an attempt to leverage the additional information highlighted by our previous experiments.
Formally, given a set of $M$ generators $g' \in \mathcal{G}$ trained on the same dataset, the cross-generator recall for a GSC trained on samples from generator $g$ is:
\begin{equation}
    r_{cross}(g) = \frac{1}{M-1}\sum_{g'\neq g \:\in \mathcal{G} } \frac{TP_g(g')}{TP_g(g')+FN_g(g')}
\end{equation}
where $TP_g(g')$ and $FN_g(g')$ denote the true positives and false negatives (considering fake as the positive class) obtained by the GSC trained on samples from generator $g$ and tested on samples from generator $g'$.

A simple aggregation of the saliency component and cross-generator recall can be:
\begin{equation}
\mathrm{CrossScore}(g)
=
s(g)
+
\mathbf{1}_{\{r_{\text{cross}}(g) > \tau\}}
\big( 1 + r_{\text{cross}}(g) \big),
\end{equation}
where $s(g)$ denotes the average median saliency value defined in Equation~\ref{eq:s(g)}, $\mathbf{1}_{\{\cdot\}}$ is the indicator function,
and $\tau=\frac{1}{M-1}\sum_{g'\neq g \in \mathcal{G}}r_{cross}(g')$, the average of $r_{cross}$ across generators $g'\neq g\in \mathcal{G}$ trained on the same dataset.

\begin{figure}[htbp]
    \centering
    \includegraphics[width=0.9\linewidth]{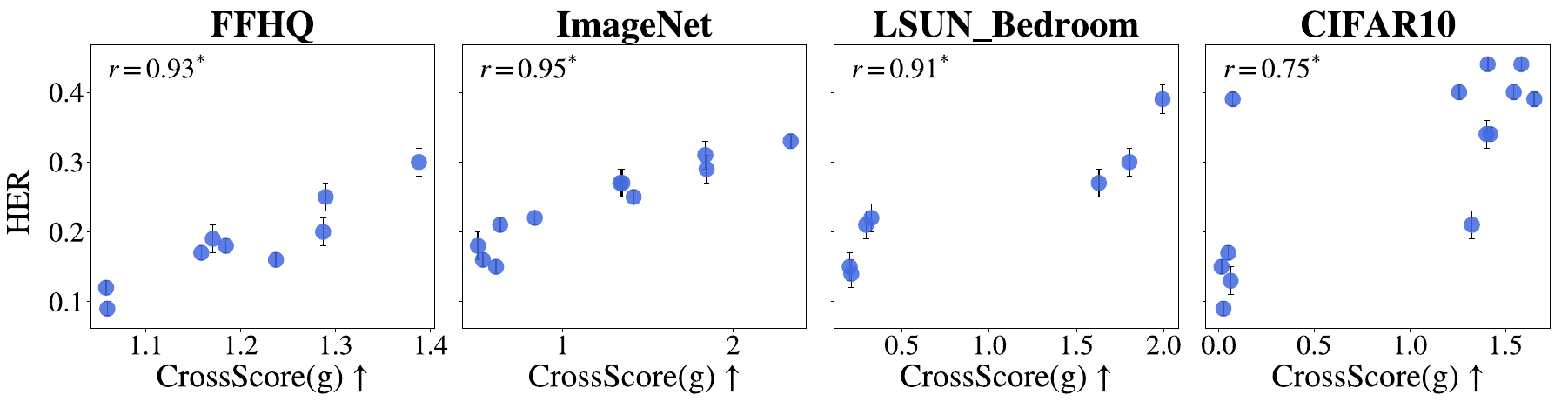}
    \caption{CrossScore vs HER. Pearson $r$ is always higher than $0.75$ and significant ($p \leq 0.05$).}
    \label{fig:ood_results}
\end{figure}

Figure~\ref{fig:ood_results} compares the proposed score with HER,
demonstrating a strong alignment with human judgments across all datasets.

When such additional information is available, the resulting score exhibits
an even more pronounced correlation with human evaluations compared to the previous formulation (note that CrossScore correlates with HER, while SAS and FID anticorrelate).
However, CrossScore is less practical as it requires access to samples from multiple alternative generators trained on the same dataset, thus it is dependent on the set of generators $\mathcal{G}$. 
Hence, we focus on SAS as our default metric: it only requires real images and samples from the 
evaluated generator, like FID.

\subsection{Validating the Expressiveness of GSC Features}

Previous works have shown that feature extractors derived from generic tasks are not naturally tailored for evaluating generative models. For the case of classification, for example, their feature spaces are optimized to separate semantic categories associated with the highest logits, rather than to capture subtle artifacts relevant to generative realism, as noted in \cite{DBLP:conf/iclr/KynkaanniemiKAA23}. For example, in the FFHQ setting, the traditional FID computed with Inception-V3 trained on ImageNet often emphasizes regions outside the face, since ImageNet does not include a dedicated “person” or “face” category. However, the objective in FFHQ is precisely to generate realistic human faces, making such features suboptimal for assessing perceptual realism.
Our goal is to evaluate the conjecture that features explicitly learned to distinguish real images from generator-specific fake ones are inherently more suitable for assessing generative realism than features learned from models trained on unrelated tasks. To test this hypothesis, we used our GSCs
as a feature extractor to compute both the Fréchet Distance and MMD.

For the Fréchet Distance,, we model the feature distributions of real samples $D_R$ and fake samples $D_F^{(g)}$ as multivariate Gaussians with empirical means $\mu_R, \mu_F^{(g)}$ and covariances $\Sigma_R, \Sigma_F^{(g)}$, computed in the feature space induced by the GSC trained for generator $g$. The resulting Fréchet Distance is defined as:
\begin{equation}
    \text{FD}_{\text{GSC}}^{(g)} = ||\mu_R-\mu_F^{(g)}||_2^{2} + Tr(\Sigma_R+\Sigma_F^{(g)}-2(\Sigma_R\Sigma_F^{(g)})^{1/2}).
\end{equation}

Because the Gaussian assumption underlying the Fréchet Distance may be
inappropriate for neural feature representations, we also evaluated the
non-parametric MMD, following the motivation of CMMD
\cite{DBLP:conf/cvpr/JayasumanaRVGCK24}. Given
$\phi_g(x)$ the feature representation of image $x$ obtained from the
GSC trained for generator $g$, the squared MMD
between the corresponding feature distributions is defined as
\begin{equation}
\begin{split}
    \operatorname{MMD}^2(D_R,D_F^{(g)})
    ={}&
    \mathbb{E}_{x,x'\sim D_R}
    \left[k\left(\phi_g(x),\phi_g(x')\right)\right] \\
    &+
    \mathbb{E}_{y,y'\sim D_F^{(g)}}
    \left[k\left(\phi_g(y),\phi_g(y')\right)\right] \\
    &-
    2\mathbb{E}_{x\sim D_R,\,y\sim D_F^{(g)}}
    \left[k\left(\phi_g(x),\phi_g(y)\right)\right],
\end{split}
\end{equation}
where $k(\cdot,\cdot)$ is a positive-definite kernel. Following \cite{DBLP:conf/cvpr/JayasumanaRVGCK24}, we use the Gaussian RBF kernel $k(u,v) =
    \exp\left(-{\lVert u-v\rVert^2}/{2\sigma^2}\right),$
with bandwidth $\sigma=10$. In practice, the expectations are replaced by their
empirical estimates over the available real and generated samples. 

\begin{table}[t]
    \centering
    \setlength{\tabcolsep}{3pt}
    \caption{Pearson correlation coefficient between HER and different evaluation
    metrics: the traditional FID
    \cite{DBLP:conf/nips/HeuselRUNH17}, Fréchet Distance and MMD computed with
    DINOv2 features, the CMMD score computed with CLIP features
    \cite{DBLP:conf/cvpr/JayasumanaRVGCK24}, and the corresponding Fréchet Distance
    and MMD scores obtained using GSCs as feature extractors. * indicates significant correlations ($p \leq 0.05$).}
    \begin{tabular}{lcccc}
    \toprule
     & \shortstack{\textbf{LSUN}\\\textbf{Bedroom}}
     & \textbf{FFHQ} 
     & \textbf{ImageNet} 
     & \textbf{CIFAR10} \\
    \midrule
    FID       & -0.11$\;\;$ & -0.59 &  0.37$\:$ & -0.96* \\
    $\text{FD}_{\text{DINOv2}}$ & -0.94* & -0.50 & -0.76* & -0.98* \\
    $\text{FD}_{\text{GSC}}$    & -0.77* & -0.49 & -0.32$\;\;$ & -0.83* \\
    CMMD & -0.98* & -0.20 & -0.67* & -0.96* \\
    $\text{MMD}_{\text{DINOv2}}$ & -0.95* & -0.79* & -0.46$\;\;$ & -0.93* \\
    $\text{MMD}_{\text{GSC}}$  & -0.97* & -0.58$\;\;$ & -0.54$\;\;$ & -0.71*\\
    \bottomrule
    \end{tabular}
    \label{tab:correlation_results}
\end{table}

The results for the Fréchet Distance show that the feature spaces learned by
our GSCs are generally more informative than those extracted from Inception-V3,
despite the considerably smaller training regime of our networks (90,000
samples). In particular, $\text{FD}_{\text{GSC}}$ achieves a stronger
correlation with HER than FID on LSUN Bedroom and ImageNet, and it anti-correlates with HER consistently for all the datasets, although it does
not outperform FID across all datasets. This finding supports
our hypothesis that even a relatively small model with a considerably smaller training regime (90,000 samples), when trained specifically
for real-versus-fake discrimination, can learn representations that are
better aligned with perceptual aspects of generative quality than those
obtained from generic classification tasks. Nevertheless, the correlation
obtained with $\text{FD}_{\text{GSC}}$ remains below that of the DINOv2-based
Fréchet Distance on most datasets, consistent with the benefits of the
large-scale pretraining used by DINOv2.

The comparison between FD and MMD further indicates that neither distributional
distance is uniformly superior. 
Overall, both FD and MMD strongly depend on the feature extractor, making it
difficult to determine whether either distance is intrinsically superior.
Their relative performance also varies across datasets: MMD provides stronger
correlations in some settings, whereas the Fréchet Distance performs better in
others. Thus, both the feature representation and the choice of distributional
distance are critical for alignment with human judgments.

\section{Conclusion and Future Directions}
In this work, we introduced a novel perspective for the evaluation of generative image
models that does not rely on large pretrained feature extractors. Our
results show that the two components of SAS, namely classification accuracy
and the contents of the saliency maps, contain meaningful and complementary
information about generative realism and exhibit strong alignment with human
perceptual judgments. Our approach is domain-agnostic in the sense that the
procedure does not depend on specific image categories or a pretrained
feature space tuned to a particular setting, and can be potentially applied
to other domains, such as for example medical images or audio spectrograms.

Importantly, we observe that GSCs do not learn a universal notion of realism, but rather capture generator-specific artifacts. This limited generalization, often seen as a drawback in detection settings, becomes informative in our context: it reflects how distinctive and structured the artifacts of a given model are.
More generally, we hope this conceptual perspective can provide a foundation for a new paradigm for evaluating generative models: generator-aware, interpretable, and grounded in perceptual evidence rather than in pretrained feature spaces.

The main limitation of our work is the need to train a dedicated generator-vs-real classifier for each setting, which is acceptable to evaluate and compare trained generators, but it is likely excessive for frequent inner-loop use during generative-model development, such as hyperparameter tuning or repeated checkpoint selection, despite being only marginally more expensive than computing FD-DINOv2..  However, this design choice is intentional and stems from our objective of avoiding reliance on pretrained models, which inevitably embed biases from their pretraining data and may not be available or reliable in specialized or underrepresented domains. In this sense, our approach trades off computational cost for increased control, transparency, and applicability to settings where pretrained feature extractors are either suboptimal or unavailable.

Future research should further investigate training strategies that
explicitly control the degree of generalization of GSCs, encouraging them to
focus only on generator-specific signals. In addition, we observed the
importance of cross-generator generalization. In our preliminary results, a
predefined pool of generators is needed, but further research might leverage
this aspect to derive new and more practical evaluation metrics with strong
perceptual alignment.

\section*{Author Contribution}
E. Ristori contributed to Conceptualization, Data curation, Methodology, Software,
Validation, Writing – original draft. L.Bindini contributed to Data curation, Methodology, Supervision, Writing – original draft. M. Lippi contributed to Conceptualization,
Methodology, Supervision, Writing – review and editing. P. Frasconi contributed to Conceptualization,
Methodology, Supervision, Writing – review and editing

\section*{AI Use Disclaimer} Artificial intelligence tools were used to assist with grammar, spelling, formatting, and improving the clarity of the writing. The authors reviewed the generated suggestions and remain fully responsible for the accuracy, content, and conclusions of the manuscript.

\bibliographystyle{elsarticle-num}

\bibliography{refs}

\appendix

\section{Additional robustness experiments}

\subsection{Sensitivity to $\alpha$}
\label{app:varying_alpha}
\begin{figure}[htbp]
    \centering
    \includegraphics[width=\linewidth]{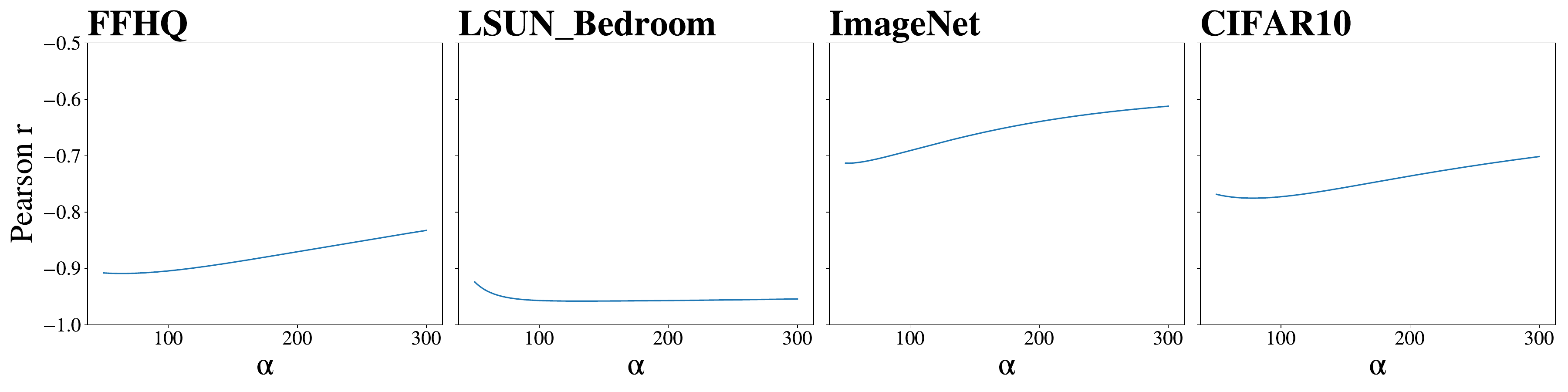}
    \caption{Pearson correlation $r$ between SAS and HER as the scaling parameter $\alpha$ varies.
Across all datasets, the correlation remains statistically significant ($p \leq 0.05$),
showing that SAS is robust to the choice of $\alpha$.}
    \label{fig:varying_alpha}
\end{figure}

We analyze the sensitivity of SAS to the scaling parameter $\alpha$.
Figure~\ref{fig:varying_alpha} reports the correlation between SAS and HER
for different values of $\alpha$.

Across datasets, the correlation remains stable and statistically significant,
indicating that the score is robust to the choice of this parameter.

\subsection{Backbone robustness}
\label{sec:performance_ViT}

\begin{figure}[htbp]
    \centering
    \includegraphics[width=0.85\linewidth]{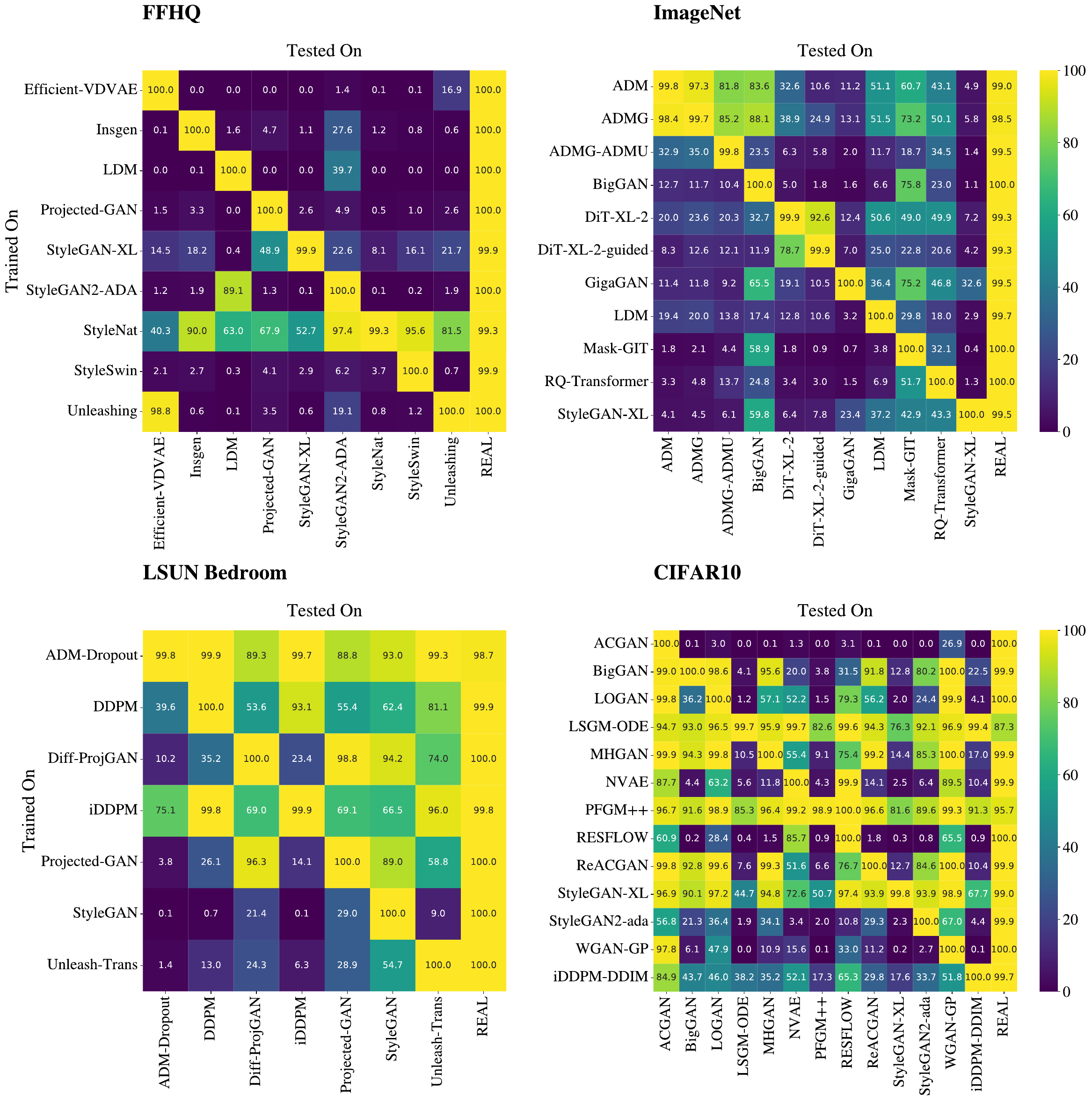}
    \caption{Cross/self generator recalls (CSGR) for ViT backbones, using a pool of $M$ generators. Specifically, the value in the $i$-th row and $j$-th column represents the cross-generator (self-generator when $i=j$) recall of the GSC trained on images from generator $i$ and tested on images from generator $j$ (with $i,j \in {1,\dots,M}$), considering fake images as the positive class. The $(M+1)$-th column shows the recall of each GSC when evaluated on real images. }
    \label{fig:confusion_matrix_ViT}
\end{figure}

We have also explored the use of GSCs based on a Vision Transformer backbone (ViT-Small) rather than a CNN to ensure the robustness of our score by varying the GSC structure and training. In Figure~\ref{fig:confusion_matrix_ViT}, we report the cross/self-generator recall (CSGR) matrices. These have the same meaning as those in Figure~\ref{fig:confusion_matrix}, but are computed using the ViT backbone.

The same qualitative patterns observed with CNNs remain visible: GSCs achieve high recall on real images and on the generator used during training, while recall decreases for samples produced by other generators.
Importantly, the cross-generator generalization behavior is preserved. Generators whose associated GSC previously exhibited stronger cross-generator recall continue to show similar behavior when trained with a ViT backbone. 

\begin{figure}[htbp]
    \centering
    \includegraphics[width=\linewidth]{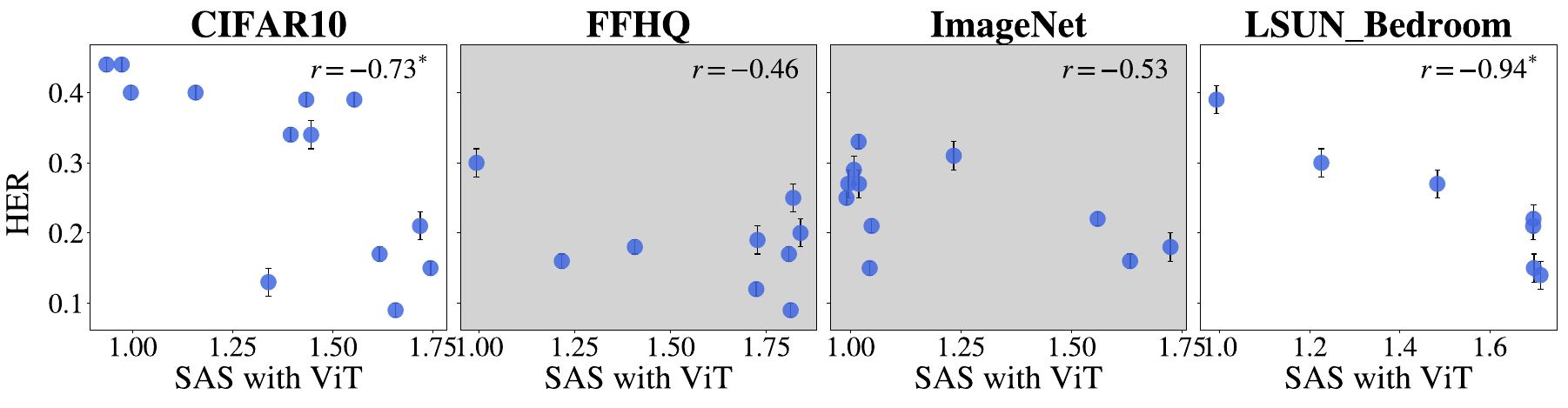}
    \caption{SAS vs HER, using a ViT as our GSC. Panels with a shaded background do not have significant ($p \leq 0.05$) correlations with HER. Compared with CNN-based classifiers (see Figure~\ref{fig:correlation_results_ours}), ViTs show weaker correlations with HER, likely because their stronger generalization and global feature representations reduce sensitivity to generator-specific artifacts.}
    \label{fig:ViT_results}
\end{figure}

In Table~\ref{tab:cross-dataset_accuracy_App} we report the cross-dataset evaluation results, corresponding to those presented for the CNN backbone in Table~\ref{tab:cross-dataset_accuracy}.
Overall, ViT-based classifiers show stronger generalization across generators and datasets compared to CNN-based GSCs, but at the same time, compared to CNN-based classifiers, the correlation with human evaluations is weaker (see Figure~\ref{fig:ViT_results}).
This is consistent with the stronger generalization ability of ViTs, which tend to capture more global and distributed image features, making them less sensitive to generator-specific artifacts.

\begin{table}[htbp]
\centering
\renewcommand{\arraystretch}{0.65}
\setlength{\tabcolsep}{2.5pt}
\caption{Cross-dataset recall for ViT-based GSCs. Classifiers trained on one dataset are evaluated on real and generated images from other datasets. \textit{Fake in} denotes the recall on fake samples generated by the same model used during training, whereas \textit{Fake out} refers to the recall on fake samples produced by generative models not seen during training.}
\begin{tabular}{llcccc}
\toprule
\multirow{2}{*}{\textbf{Training}} 
& \multirow{2}{*}{\textbf{Recall}} 
& \multicolumn{4}{c}{\textbf{Test dataset}} \\
\cmidrule(lr){3-6}
& & \textbf{LSUN} & \textbf{ImageNet} & \textbf{FFHQ} & \textbf{CIFAR10} \\
\midrule
\multirow{3}{*}{LSUN}
 &  Real                  & 99.74     & 84.21     & 74.01     & 66.19 \\
 &  Fake out              & 51.68     & 38.71     & 38.35     & 68.25 \\
 &  Fake in               & 99.84     & -         & -         & - \\
\addlinespace[2pt]
\multirow{3}{*}{ImageNet} 
 &  Real                  & 96.12     & 99.45     & 98.47     & 44.25 \\
 &  Fake out              & 30.96     & 24.26     & 17.96     & 80.94 \\
 &  Fake in               & -         & 94.93     & -         & - \\
\addlinespace[2pt]
\multirow{3}{*}{FFHQ} 
 &  Real                  & 92.27     & 95.34     & 99.86     & 58.99 \\
 &  Fake out              & 13.6      & 15.39     & 15.17     & 61.65 \\
 &  Fake in               & -         & -         & 99.49     & - \\
\addlinespace[2pt]
\multirow{3}{*}{CIFAR10} 
 &  Real                  & 100.00    & 99.71     & 99.50     & 98.78 \\
 &  Fake out              & 0.38      & 0.94      & 1.98      & 48.82 \\
 &  Fake in               & -         & -         & -         & 99.87 \\
 
\bottomrule
\end{tabular}

\label{tab:cross-dataset_accuracy_App}
\end{table}

\section{The importance of training from scratch}
\label{app:DINOv2 GSCs}
In this section, we investigate whether training a classifier from scratch is
necessary. To this end, we analyze a GSC that does not receive real and
generated images as input, but rather their DINOv2 features. The classifier is
therefore trained to distinguish real from generated images directly in the
DINOv2 feature space.

As shown in Table~\ref{tab:dino_features_GSC}, the model trained on DINOv2 features exhibits
substantially stronger generalization capabilities than the ResNet-based GSCs.
Although this behavior is desirable when the objective is fake
detection, it is less desirable when the goal is to assess the fingerprint of
a specific generative model.

\begin{table}[htbp]
\centering
\renewcommand{\arraystretch}{1.05}
\setlength{\tabcolsep}{5.0pt}
\caption{Recall for GSCs trained on DINOv2 features rather than on raw images. \textit{Fake-in} denotes the recall on fake samples generated by the same model used during training, whereas \textit{Fake-out} refers to the recall on fake samples produced by generative models not seen during training.\label{tab:dino_features_GSC}}
\begin{tabular}{lccc}
\toprule
\multirow{2}{*}{\textbf{Dataset}} 
& \multicolumn{3}{c}{\textbf{Recall}} \\
\cmidrule(lr){2-4}
& \textbf{Real} & \textbf{Fake-in} & \textbf{Fake-out}  \\
\midrule
LSUN Bedroom & 98.44 & 98.14 & 84.87 \\
ImageNet & 95.41 & 95.39 & 72.91\\
FFHQ & 98.24 & 98.51 & 77.72 \\
CIFAR10 & 94.33 & 95.36 & 70.41\\
\bottomrule
\end{tabular}
\end{table}

\begin{figure}[htbp]
    \centering
    \includegraphics[width=0.8\linewidth]{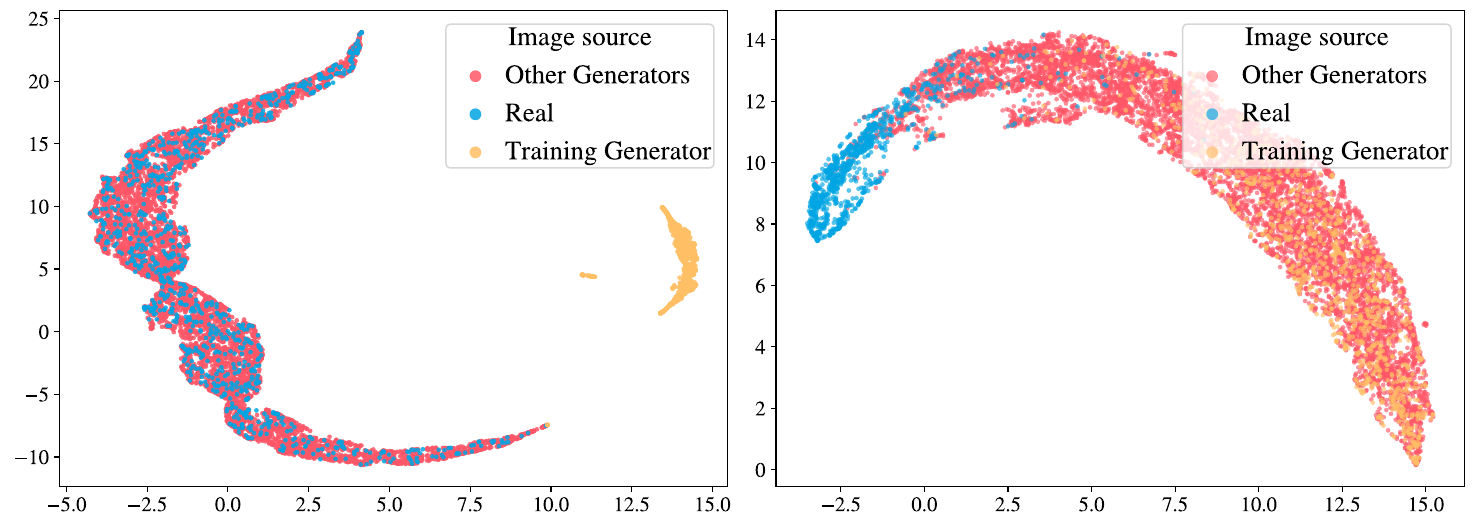}
    \caption{On the left: UMAP of the feature space of our GSC; the samples from the generator seen during training are well separated, while the other generated samples and the real ones are perfectly overlapped. On the right: UMAP projection of GSC representations learned from DINOv2 features;
    real samples occupy a largely distinct region, indicating effective
    real-versus-fake discrimination, while samples from different generators substantially overlap, suggesting limited sensitivity to generator-specific fingerprints.\label{fig:UMAP_features}}
\end{figure}

The UMAP projections in Figure~\ref{fig:UMAP_features} further illustrate the different
representations learned by GSCs trained on images and on DINOv2 features. In
the former case, features of samples generated by the model on which the GSC was trained
are separated from all other samples, while real images and samples
generated by other models appear overlapping each other. In contrast, for the classifier trained on DINOv2
features, the UMAP projection shows that real and generated samples are organized along a continuous
manifold. Nevertheless, the real samples remain concentrated in a distinct
region with little overlap with the samples produced by either the training
generator or the other generators. This separation suggests that the
DINOv2-based GSC learns features that are effective for distinguishing real
images from fake ones, including fake images generated by models that were
not used during training. At the same time, the generated samples from the
different models substantially overlap, and the samples from the training
generator are only slightly more separated from those produced by the other
generators. Thus, the representation appears well suited to generic
fake-image detection and exhibits good cross-generator generalization, but it
captures generator-specific fingerprints less distinctly than the GSC trained
directly on images.

\end{document}